\documentclass[final,5p,times,twocolumn]{elsarticle}

\usepackage{amssymb}

\usepackage{booktabs}
\usepackage[table,xcdraw]{xcolor}
\usepackage{multirow}
\usepackage{graphicx} 
\usepackage{subcaption}
\usepackage[capitalise]{cleveref}
\usepackage{url}
\usepackage{todonotes}
\usepackage{arydshln}

\begin{document}

\begin{frontmatter}

\title{Benchmarking Domain Generalization Algorithms in Computational Pathology
}

\author[inst1]{Neda Zamanitajeddin\fnref{label1}}
\author[inst1]{Mostafa Jahanifar\fnref{label1}}
\author[inst1]{Kesi Xu}
\author[inst3]{Fouzia Siraj}
\author[inst1,inst2]{Nasir Rajpoot \corref{corrLabel}}

\fntext[label1]{Joint first authors, contributed equally.}
\cortext[corrLabel]{Corresponding authors: n.m.rajpoot@warwick.ac.uk}
\affiliation[inst1]{organization={Tissue Image Analytics centre, Department of Computer Science},
            addressline={University of Warwick}, 
            city={Coventry},
            postcode={UK}}
\affiliation[inst2]{organization={Histofy Ltd},
            city={Coventry},
            postcode={UK}}
\affiliation[inst3]{organization={ICMR-National Institute of Pathology},
            city={New Delhi, India}
            }

\begin{abstract}
Deep learning models have shown immense promise in computational pathology (CPath) tasks, but their performance often suffers when applied to unseen data due to domain shifts. Addressing this requires domain generalization (DG) algorithms. However, a systematic evaluation of DG algorithms in the CPath context is lacking. This study aims to benchmark the effectiveness of 30 DG algorithms on 3 CPath tasks of varying difficulty through 7,560 cross-validation runs. We evaluate these algorithms using a unified and robust platform, incorporating modality-specific techniques and recent advances like pretrained foundation models. Our extensive cross-validation experiments provide insights into the relative performance of various DG strategies. We observe that self-supervised learning and stain augmentation consistently outperform other methods, highlighting the potential of pretrained models and data augmentation. Furthermore, we introduce a new pan-cancer tumor detection dataset (HISTOPANTUM) as a benchmark for future research. This study offers valuable guidance to researchers in selecting appropriate DG approaches for CPath tasks.
\end{abstract}


\end{frontmatter}


\section{Introduction}
\label{sec:intro}

\begin{figure*}
    \centering
    \includegraphics[width=1\linewidth]{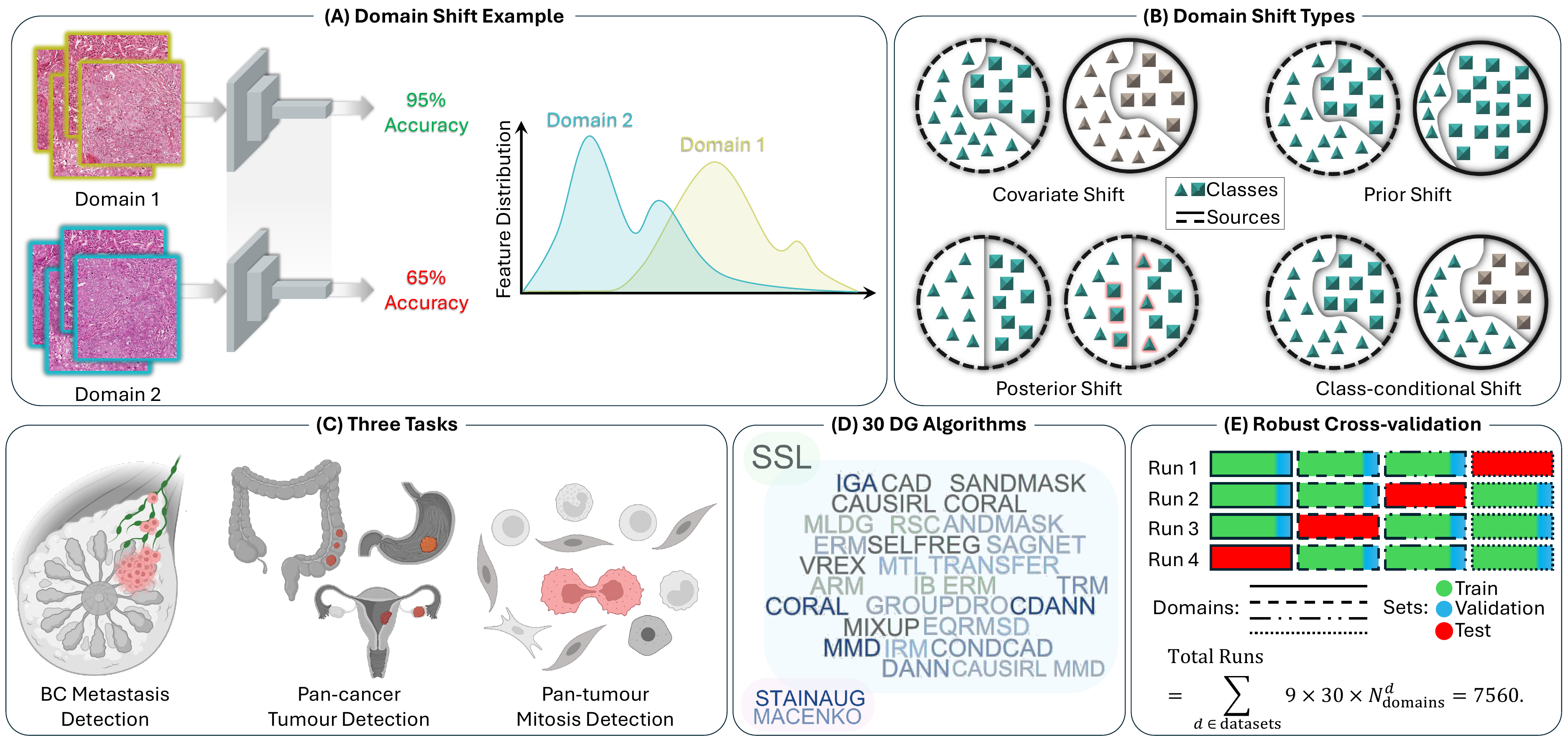}
    \caption{Domain shift in computational pathology can cause degradation in performance when testing on an unseen dataset (A). Different types of DS are illustrated in (B) with shapes as classes, colors as features, and each circle as a domain. In (B), covariate shift is presented by changing the color of objects in two different domains, prior shift happens when the distribution of classes differs between the two domains, and posterior shift is shown when the same objects are labeled differently by the observers (highlighted shapes), and in class-conditional shift, the color of only one class is changing between domains. Leveraging three different tasks (C) in this work, we benchmark the performance of 30 domain generalization algorithms (D) in a series of robust cross-validation experiments (E).}
    \label{fig:overview}
\end{figure*}

Deep learning (DL) models have shown great potential in addressing fundamental problems in computational pathology (CPath) \cite{litjens2017survey,van2021deep} such as histology image classification \cite{bilal2023development,graham2023screening,180_vuong2023cpath}, tissue segmentation \cite{119_graham2023cpath,jahanifar2021robust,shephard2021simultaneous}, and nuclei detection \cite{graham2023conic,59_alemi-koohbanani2020cpath,jahanifar2024mitosis}. Furthermore, leveraging DL, more advanced problems are being tackled in the field, such as gene expression prediction \cite{dawood2023cross,zamanitajeddin2024social,bilal2021development} and biomarker discovery \cite{lu2024ai,wahab2023ai,ibrahim2024artificial}. Regardless of the versatility and accuracy of DL models on the training domain data, it has also been shown that testing on unseen data can degrade the performance metrics considerably \cite{jahanifar2023domain} (see \cref{fig:overview}A) which is a phenomenon usually caused by domain shift (DS) \cite{13_stacke2021cpath}.

Defining a data `domain' as the joint distribution of feature ($X$) and label ($Y$) spaces, domain shift can be characterized by discrepancies in the joint distribution of features and labels across source ($s$) and target ($t$) domains, that is, $P^s_{XY} \neq P^t_{XY}$. According to the Bayes' theorem, the joint distribution is given by ${P_{XY}} = {P_{X|Y}}{P_Y} = {P_{Y|X}}{P_X}$ which can be leveraged to describe DS manifested in various ways \cite{jahanifar2023domain}:
\begin{itemize}
    \item \textbf{Covariate Shift:} The feature distributions differ between the source and target domains, i.e., $P^s_X \neq P^t_X$. \textit{Example:} Tissue samples scanned using different scanners exhibit distinct colors and features.
    
    \item \textbf{Prior Shift:} The label distributions vary between the source and target domains, i.e., $P^s_Y \neq P^t_Y$. \textit{Example:} A model trained on a dataset with a specific proportion of cancerous to non-cancerous samples is applied to a new dataset with a different ratio of these classes.
    
    \item \textbf{Posterior Shift:} The conditional label distributions differ, i.e., $P^s_{Y|X} \neq P^t_{Y|X}$. \textit{Example:} Subjective labeling in mitosis detection where different annotators assign different labels to the same data due to varying interpretations.
    
    \item \textbf{Class-Conditional Shift:} The data characteristics for a specific class differ between the source and target domains, i.e., $P^s_{X|Y=y} \neq P^t_{X|Y=y}$. \textit{Example:} Morphological traits of cancer cells in early-stage cancers differ from those in late-stage cancers, leading to variations in the same class across datasets.
\end{itemize}

A schematic presentation of different DS types is adopted from \cite{jahanifar2023domain} and shown in \cref{fig:overview}B.

An ideal way of, addressing domain shifts would involve training models across all conceivable data distributions. However, this approach is typically infeasible due to the limited availability of comprehensive, multi-domain data during the training phase. Consequently, there is an urgent need for algorithms specifically designed to improve domain generalization (DG). DG refers to the capability of a model trained on data from source domains \(D^s\) to perform well on unseen target domains \(D^t\) despite distributional differences (\(P^s_{XY} \neq P^t_{XY}\)). It is important to note that, unlike domain adaptation techniques \cite{12_guan2022general}, DG algorithms enhance generalization to novel target domains without access to target domain data during training \cite{112_wang2021general,jahanifar2023domain}.

There remains a critical gap in the utilization of existing DG algorithms within CPath. Many sophisticated DG algorithms have not been systematically explored in this field. Motivated by this gap, we present a rigorous evaluation of DG methods in the CPath context. This research aims to benchmark the effectiveness of 30 different DG algorithms (\cref{fig:overview}D) on three different CPath tasks of various difficulties (\cref{fig:overview}C) in a unified and robust platform through extensive cross-validation experiments (\cref{fig:overview}E). Our goal is to fairly compare existing  DG algorithms to provide insights that could help researchers to select a better strategy for their DG needs. To this end, we build on the DomainBed platform \cite{domainbed} as our main repository of DG algorithms, by including new DG algorithms (CPath-specific algorithms such as stain normalization and stain augmentation, and more recent approaches such as a pretrained foundation model), more robust evaluation metric (F1 score), and two new multi-domain histology image classification datasets. 
This work's main contributions can be summarized as follows:
\begin{itemize}
    \item Presentation of a unified and robust framework for benchmarking DG algorithms in the area of CPath;
    \item Benchmarking 30 DG algorithms on 3 different tasks in CPath using the proposed framework;
    \item Comprehensive and robust cross-validation experiments, covering 7,560 training-validation runs;
    \item Making recommendations for selecting effective DG strategies in CPath; and
    \item Releasing a large-scale tumor patch dataset (that we term HISTOPANTUM) comprising 280K+ images in 4 different cancer types and capturing three types of DG and making the benchmarking framework \textit{HistoDomainBed}  publicly available at: \url{https://github.com/mostafajahanifar/HistoDomainBed}.
\end{itemize}

In the remainder of the paper, datasets, algorithms, and cross-validation procedure are described in \cref{sec:meto}, results are presented and discussed in sections \ref{sec:res} and \ref{sec:disc}, and finally, paper is concluded in \cref{sec:conc}.

\section{Material and Methods}
\label{sec:meto}

\subsection{Datasets and tasks}
\label{sec:data}

\begin{figure*}[!t]
    \centering
    \includegraphics[width=\textwidth]{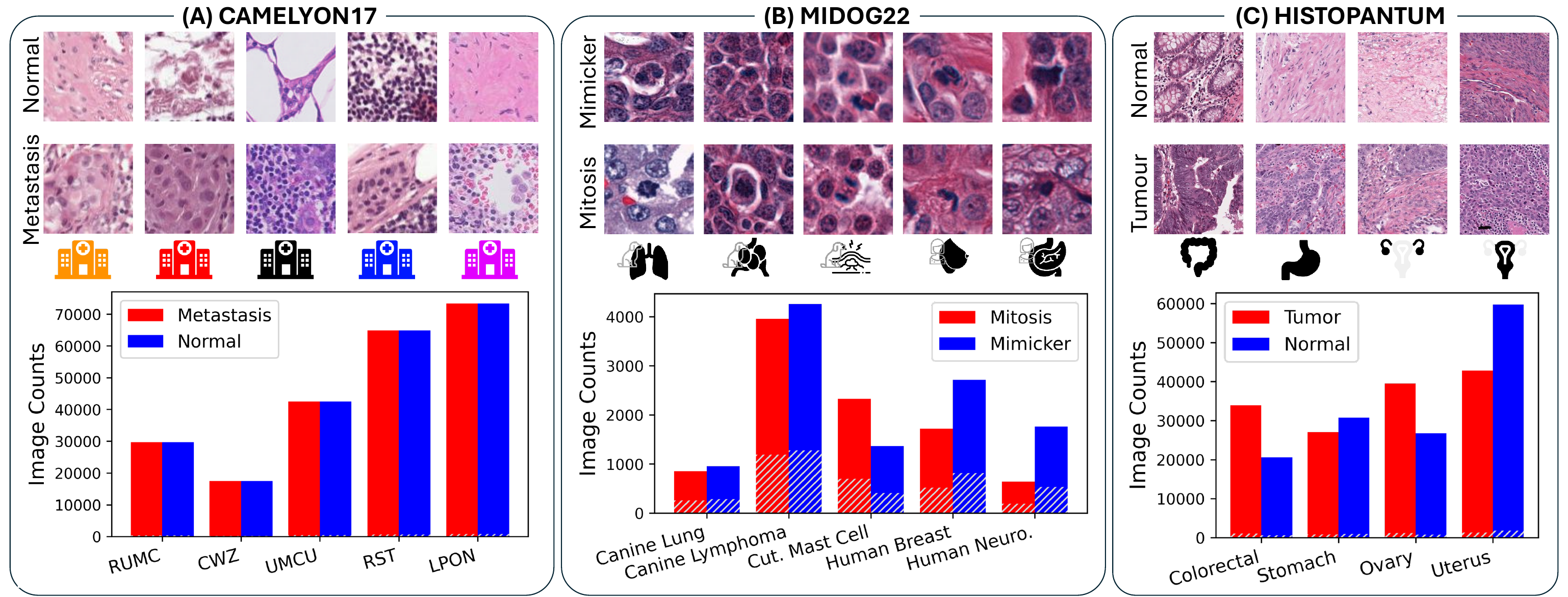}
    \caption{Tasks and datasets used in the benchmarking process: (A) Breast cancer metastasis detection leveraging Camleyon17 dataset \cite{75_bandi2019cpath}, (B) Mitosis detection in MIDOG22  dataset \cite{aubreville2024domain}, and (C) tumor detection in our proposed HISTOPANTUM dataset. For every dataset, an example from each domain and class is provided. All the tasks are designed as a binary classification task, where the name and population of positive and negative classes are shown in red and blue color bars, respectively. The hatched region in each bar represents the fraction of samples used to generate small datasets (see \cref{sec:sub-dataset}).}
    \label{fig:data}
\end{figure*}

In this study, we examine three datasets — CAMELYON17 \cite{75_bandi2019cpath}, MIDOG22  \cite{aubreville2024domain}, and HISTOPANTUM — each chosen for its specific challenges and domain shifts to facilitate a comprehensive evaluation of DG algorithms applied to various classification tasks in CPath.

\subsubsection{CAMELYON17}

With the CAMELYON17 grand challenge on the detection of breast cancer metastases in sentinel lymph nodes, a dataset with the same name has been released which comprises Whole Slide Images (WSIs) of lymph node resections in breast cancer patients and their corresponding lesion-level annotations \cite{75_bandi2019cpath}. The provided annotations have been leveraged for the extraction of a patch-level dataset of metastatic and normal images of breast cancer. In particular, we used the CAMELYON17 dataset part of the WILDS toolbox \cite{137_koh2021cpath} (which is designed to test machine-learning models against significant distribution shifts) to allow for the reproducibility of the results. 

CAMELYON17 is gathered from various medical centers in the Netherlands, including Radboud University Medical Center (RUMC), Canisius-Wilhelmina Hospital (CWZ), University Medical Center Utrecht (UMCU), Rijnstate Hospital in Arnhem (RST), and the Laboratory of Pathology East-Netherlands (LPON), each representing a unique domain. CAMELYON17 comprises 455,953 image patches, categorized into metastasis and non-metastasis (normal) classes, with each patch measuring $96\times96$ pixels at a resolution of 0.5 microns per pixel (mpp).

A major challenge presented by CAMELYON17 is the covariate shift, primarily caused by variations in imaging equipment and procedures across different centers. These variations are evident in the noticeable differences in color and texture among images from various sources as shown in \cref{fig:data}A.
Despite covariate shift, the dataset is well balanced in label space, with an equal number of tumor and non-tumor patches in each domain, effectively eliminating any potential prior shift in label distribution. The objective nature of the classification task ensures that there is no posterior shift, focusing the analysis solely on addressing the implications of covariate shifts.

\subsubsection{MIDOG22}
\label{sec:MIDOG22}
The second task, mitosis detection, involves classifying mitotic figures versus mimickers (cells of other types that are very similar to mitotic figures in appearance), which is a binary classification task previously explored in the literature \cite{jahanifar2023domain,jahanifar2024mitosis,192_kotte2022cpath,61_wang2023cpath}.

For this purpose, we utilize the MIDOG22  dataset \cite{aubreville2024domain}, which comprises five domains: Canine Lung Cancer, Human Breast Cancer, Canine Lymphoma, Canine Cutaneous Mast Cell Tumor, and Human Neuroendocrine Tumor. From the original MIDOG22  dataset and based on the annotations provided, we extract 20,552 image patches, each sized $128\times128$ pixels at a resolution of 0.25 mpp, and categorize them into mitosis and mimicked classes.

MIDOG22  is particularly challenging for achieving DG due to the presence of all four types of DS: covariate shift is evident as the images come from different centers using various scanners, leading to variations in color schemes. There is a prior shift, as different labels are distributed variably across domains, clearly shown in the dataset \cref{fig:data}B.
The task also involves a posterior shift due to the highly subjective nature of mitosis labeling. Additionally, class-conditional shifts occur because different tumor types and species influence the appearance of non-mitotic regions in the images, significantly varying from one domain to another.
The complexity of this dataset makes it a rigorous test bed for assessing the DG capabilities of different algorithms in CPath.

\subsubsection{HISTOPANTUM}
The last task we address is pan-cancer tumor detection, leveraging the HISTOPANTUM dataset that we were releasing in this study. This dataset captures four different cancer types: Colorectal (CRC), Uterus (UCEC), Ovary (OV), and Stomach (STAD), collectively referred to as four domains. During data curation, we source 40 WSIs for each cancer type from its related study in The Cancer Genome Atlas Program (TCGA) \cite{168_weinstein2013cpath}. For sampling, we make sure to include a variety of tumor subtypes (adenocarcinoma and mucinous carcinoma), genders, ethnicities, and centers in order to make the dataset as diverse as possible. Then, an experienced pathologist (FS) meticulously annotates tumor and non-tumor regions in the slides, which are used to extract tumor and non-tumor patches from the WSIs to form the HISTOPANTUM dataset.

The HISTOPANTUM dataset includes 281,142 patches, each $512\times512$ pixels at a resolution of about 0.5 mpp which are subsequently resized to $224\times224$ pixels during training and evaluation. HISTOPANTUM patches are classified into two classes: tumor and non-tumor (normal). 
This dataset presents three significant types of DS. Firstly, a covariate shift arises due to images being sourced from different centers using different slide preparation and scanning devices, introducing notable variations in color and stain schemes. Secondly, a prior shift is observed with a distinct distribution of classes across domains, as depicted in \cref{fig:data}C. Lastly, the class-conditional shift is evident as different cancer types influence the appearance of  tumor region in the images (morphology of tumor cells varies significantly from one tumor to another), while non-tumor regions remain relatively consistent across domains (for example, the morphology of stromal and inflammatory regions are very similar across different cancer types). 
Notably, this dataset does not suffer from the posterior shift, thanks to the objective nature of the labeling process, eliminating subjectivity in tumor detection. We are making the HISTOPANTUM publicly available as a benchmark for pan-cancer tumor detection \footnote{Data is being uploaded to a web server. Please check this link for updates: \url{https://github.com/mostafajahanifar/HistoDomainBed}}.

\subsubsection{Subsampled datasets}
\label{sec:sub-dataset}

In addition to the primary experiments, we conduct a series of tests to understand how different DG algorithms perform under a low-data budget scenario. This analysis is crucial for applications where data availability is limited. For this purpose, we create smaller versions of the original datasets, maintaining similar distributions but significantly smaller populations.

The reduced (small) datasets are generated by randomly sampling the following percentages of each class in the original datasets (as hatched regions shown in the bar plots of \cref{fig:data}):
\begin{itemize}
    \item \textbf{sCAMELYON17}: 1\% of the original CAMELYON17 dataset (N=4,560).
    \item \textbf{sHISTOPANTUM}: 3\% of the original HISTOPANTUM dataset (N=8,434).
    \item \textbf{sMIDOG22 }: 30\% of the original MIDOG22  dataset (N=6,166).
\end{itemize}

The sampling percentage in each dataset is set to considerably reduce dataset size while keeping enough samples for convergence of DG algorithms. The same experimental setup, model selection strategy, and algorithms used in the original experiments are applied to these smaller datasets. This approach allows us to directly compare the performance and generalization capability of the algorithms in both `large dataset' and `small dataset' scenarios.

\subsection{Algorithms}

\begin{figure}[t]
    \centering
    \includegraphics[width=0.7\columnwidth]{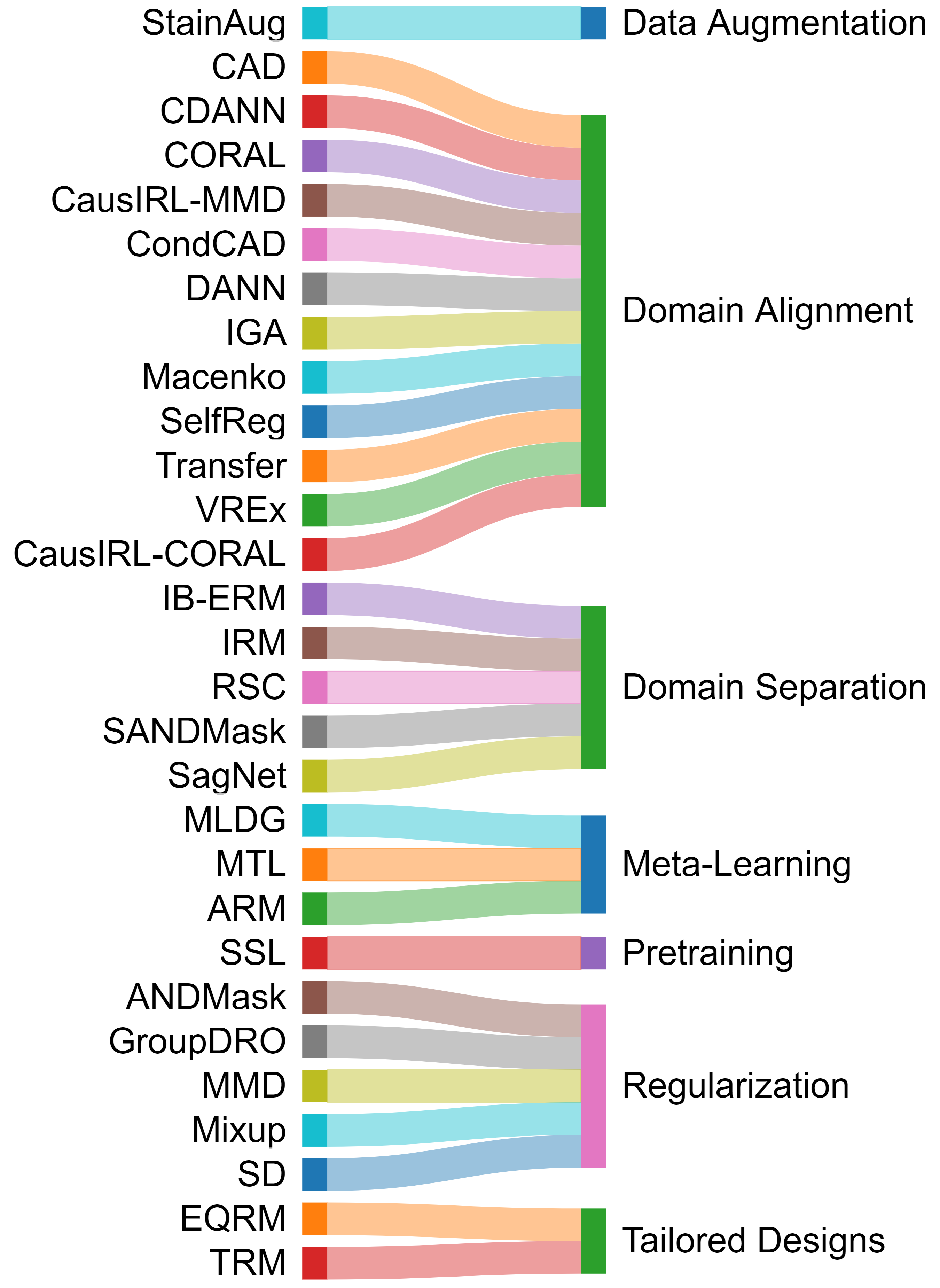}
    \caption{Benchmarked algorithms categorized into different domain generalization methodologies, as introduced in \cite{jahanifar2023domain}}
    \label{fig:algo}
\end{figure}

We utilize DomainBed \cite{domainbed} as our main benchmarking tool because it offers a robust and well-tested platform for fair and reproducible comparison of different DG algorithms. Furthermore, DomainBed is well maintained and contains the most number of state-of-the-art DG algorithms implemented in comparison to other DG benchmarking tools (such as DeepDG \cite{deepdg}). Using DomainBed, we can test different DG algorithms while also controlling algorithms and training hyperparameters. As well as the DomainBed implemented algorithms, we also investigate the two CPath-specific algorithms, namely stain normalization and stain augmentation, and a self-supervised learning (SSL) based algorithm in the same structured experimental setup of DomainBed. Furthermore, we have also added the F1-score evaluation metric to the platform which originally included only the Accuracy metric. The code base for our updated DomainBed platform, called \textit{HistoDomainBed}, is available at: \url{https://github.com/mostafajahanifar/HistoDomainBed}.

All the algorithms in our experiments use a standard ResNet50 \cite{he2016deep} model for feature extraction due to its proven generalization capabilities and popularity. The preferred model selection strategy in our work is the ``training-domain validation set'', recognized for its effectiveness in different scenarios and datasets as shown previously in \cite{domainbed}. More information on how we performed cross-validation experiments using DomainBed is given in \cref{sec:cross-val}. 

In the rest of this section, we introduce the DG algorithms investigated in this work. Explaining the methodology of each algorithm is outside of the scope of this work, although we have categorized these algorithms into 6 distinct categories of DG methods based on their working principles and the introduction of the categories in \cite{jahanifar2023domain}.
\textit{Domain alignment} techniques bridge domain gaps by harmonizing feature representations, employing methods like stain normalization and generative models.
\textit{Data augmentation} enhances model generalization through image transformations and generative networks.
\textit{Meta-learning} enables quick adaptation to new domains or tasks using techniques like MAML \cite{finn2017maml}.
\textit{Tailored model design} strategies leverage the unique characteristics of histopathology images with specialized network architectures and loss functions. \textit{Pretraining} strategies use self-supervised, unsupervised, and semi-supervised learning to enhance feature encoding and generalizability. \textit{Regularization} strategies prevent overfitting and improve performance on unseen data by introducing constraints and penalties.
\textit{Domain separation} learns disentangled domain-specific and domain-agnostic features. 
 
A chart showing the investigated DG algorithms and their related category is presented in \cref{fig:algo}. For more information on these categories or related methods, please refer to \cite{jahanifar2023domain} or the respective cited articles.


\subsubsection{DomainBed algorithms}
DomainBed \cite{domainbed}, developed by the Facebook Research group, is a comprehensive PyTorch suite. At the time of writing this manuscript, it encompassed support for 27 DG algorithms, 10 computer vision datasets, and one CPath dataset (Camelyon17 WILDS \cite{137_koh2021cpath}). The toolkit includes algorithms such as Empirical Risk Minimization (ERM) \cite{vapnik1999erm}, Interdomain Mixup (Mixup) \cite{yan2020mixup}, Group Distributionally Robust Optimization (GroupDRO) \cite{sagawa2019groupdro}, Conditional Domain Adversarial Neural Network (CDANN) \cite{135_li2018general}, Learning Explanations that are Hard to Vary (AND-Mask) \cite{parascandolo2020andmask}, Deep CORAL (CORAL) \cite{sun2016deepcoral}, Self-supervised Contrastive Regularization (SelfReg) \cite{kim2021selfreg}, Marginal Transfer Learning (MATL) \cite{blanchard2021mtl}, and Adaptive Risk Minimization (ARM) \cite{zhang2021arm}. It also supports Invariant Risk Minimization (IRM) \cite{arjovsky2019irm}, Domain Adversarial Neural Network (DANN) \cite{ganin2016dann}, Style Agnostic Networks (SagNet) \cite{nam2021sagnet}, Learning Representations that Support Robust Transfer of Predictors (TRM) \cite{xu2021trm}, Optimal Representations for Covariate Shift (CAD and CondCAD) \cite{dubois2021cad}, Representation Self-Challenging (RSC) \cite{huang2020rsc}, Maximum Mean Discrepancy (MMD) \cite{li2018mmd}, Out-of-Distribution generalization with Maximal Invariant Predictor (IGA) \cite{koyama2020iga}, Variance Risk Extrapolation (VREx) \cite{krueger2021vrex}, and Invariance Principle Meets Information Bottleneck for Out-of-Distribution generalization (IB-ERM) \cite{ahuja2021ib}. Additional algorithms include Empirical Quantile Risk Minimization (EQRM) \cite{eastwood2022eqrm}, Spectral Decoupling (SD) \cite{pezeshki2021sd}, Quantifying and Improving Transferability in Domain generalization (Transfer) \cite{zhang2021transfer}, Smoothed-AND mask (SAND-mask) \cite{shahtalebi2021sand}, Meta-Learning Domain generalization (MLDG) \cite{li2018mldg}, and Invariant Causal Mechanisms through Distribution Matching (CausIRL with CORAL or MMD) \cite{chevalley2022CausIRL}.

\subsubsection{CPath-specific algorithms}

Ideally, the same tissue specimens, stained in different laboratories, should yield identical results, but this ideal is often unattainable. Stain variation can arise from differences in slide scanners, stain quality and concentration, and staining procedure \cite{197_bejnordi2014cpath}. Pathologists can easily disregard irrelevant features (such as stain variation) in a WSI that do not impact their diagnosis. However, deep learning models sometimes struggle with this task \cite{21_tellez2019cpath}. 

\paragraph{Stain Normalization} A preprocessing step that aligns the color distributions of histology images to a reference image, countering discrepancies from varied staining procedures and scanners. Stain normalization techniques range from linear scaling and histogram matching to advanced methods by Ruifrok \cite{198_ruifrok2001cpath}, Macenko \cite{199_macenko2009cpath}, and Vahadane \cite{201_vahadane2016cpath}. These methods adjust the stain matrix of source images while maintaining their stain concentrations, ensuring color consistency without altering structural details. In this work, we compare Macenko \cite{199_macenko2009cpath} stain normalization algorithm implemented in TIAToolbox \cite{pocock2022tiatoolbox}. To this end, we normalize the stain of all images in all datasets offline, utilizing the same reference image, to maintain training efficiency. 

\paragraph{Stain Augmentation (StainAug)} A method that involves decomposing RGB histology images into stain components, perturbing them, and recomposing the images to introduce variability. This technique has proven effective in numerous tasks and helps improve model generalizability \cite{jahanifar2023domain, 21_tellez2019cpath,faryna2024automatic}. In HistoDomainBed, stain augmentation was randomly done on the fly using TIAToolbox \cite{pocock2022tiatoolbox}. To this end, the Macenko method \cite{199_macenko2009cpath} was used to extract the stain matrix and components from the RGB image.


It is important to note that both Macenko and StainAug algorithms are employed on top of the Empirical Risk Minimization (ERM) \cite{vapnik1999erm} approach. In other words, Macenko refers to the scenario where we use ERM on stain-normalized images and StainAug is the scenario where the stain augmentation technique is used during the training of the ERM model on original images.

\subsubsection{Self-supervised learning (SSL)}
For our SSL experiments, we use a ResNet50 model pretrained on histology images to start each run instead of initializing the model with ImageNet weights. In particular, we choose a foundation model from the work of Kang \textit{et al.} \cite{159_kang2023cpath} which has a ResNet50 architecture and is pretrained on 19M image patches from different studies of TCGA using Barlow Twin self-supervised learning algorithm \cite{zbontar2021barlow}. The utilized SSL algorithm also works on top of the ERM algorithm in our HistoDomainBed platform. 

\subsection{Cross-validation and model selection}
\label{sec:cross-val}

Gulrajani \textit{et al.} \cite{domainbed} meticulously examined three distinct model selection scenarios, which are crucial for determining how the performance of a model on a validation set can guide the selection of the best training epoch for application on unseen test sets. The scenarios we explore are:

\begin{enumerate}
    \item Training-domain validation set: This involves pooling a validation set from all training domains, a standard practice of training/validation split across domains. The best-performing model of the validation set is then tested on hold-out test domains.
    \item Leave-one-domain-out validation: In this scenario, one of the training domains is reserved for validation. The model that performs best on this holdout domain is then re-trained on all domains before being applied to the test domains.
    \item Test-domain validation set (oracle): This scenario selects the model that maximizes accuracy on a validation set mirroring the test domain's distribution, albeit with limited queries. Although this is theoretically not a valid model selection method due to its reliance on access to the test domain, in the original DomainBed \cite{domainbed} it was included for comparative analysis.
\end{enumerate}

Based on findings of the DomainBed study \cite{domainbed}, the ``training-domain validation set" model selection scenario consistently yielded the best results on different datasets and using different algorithms. Thus, we have chosen to adopt this model selection strategy for our experiments to allow us to minimize computational demands and focus our resources on evaluating the effectiveness of the DG algorithms themselves, rather than delving into various model selection techniques.
 In our model selection strategy, we allocate 20\% of the training data for validation purposes.
 
 For robust cross-validation, one domain was systematically left out for testing, and this process was repeated across the number of domains available in each task (Run 1, 2, 3, ...). An illustration of the utilized cross-validation process is given in \cref{fig:overview}E, where there are 4 domains and therefore 4 runs, and in each run data from three domains are used for training and validation (80\%-20\%), and 1 domain is left out for testing. All the metrics reported in this work are based on results of experiments on unseen test domains. 

To ensure the reliability of our results, we varied the hyperparameters (such as batch size, learning rate, and algorithm-specific parameters) randomly three times for each experiment. Each set of hyperparameters was then used to conduct three independent runs, leading to a total of nine training runs for each domain and method combination. The working range of algorithms' hyperparameters is selected based on convergence experiments done for each algorithm beforehand.
In summary, this comprehensive cross-validation approach resulted in a total of 7,560  training-validation runs for both full and small datasets ($\sum\nolimits_{d \in {\rm{datasets}}} {9 \times 30 \times N_{{\rm{domains}}}^d}$), illustrating the extensive scale and rigorous nature of our experimental design.

In all the runs, we utilize an ImageNet-pretrained ResNet50 model \cite{he2016deep} (except for SSL algorithm which uses histology-pretrained weights as a starting point), trained for approximately 30 epochs using an Adam optimizer. All the experiments are performed using 8 NVidia Tesla V100 GPUs on a DGX2 machine.

\section{Results}
\label{sec:res}

This section presents the performance metrics of different algorithms across various datasets. The metrics reported are binary F1 score and accuracy. We added the F1 score to HistoDomainBed to account for the scenarios where the data is significantly imbalanced, rendering accuracy a sub-optimal metric. Using Accuracy and F1 Score, we can comprehensively understand our models’ performance across different domains, ensuring that our evaluation is robust and reliable.

\subsection{Results for full datasets}

We present the performance metrics of various algorithms across full-scale datasets in \cref{tab:full_res}. Accuracy and F1 metrics are reported for each dataset (task) separately as well as for the average performance across all the tasks. The rows in \cref{tab:full_res} are ordered by the average F1 score across all tasks. In each column, cells are colored from red to green based on the performance values, red indicating worse and green indicating better performance.

The majority of methods exhibited similar performance, with average F1 scores ranging from 81\% to 85\%, except for the top 2 algorithms. SSL \cite{159_kang2023cpath} and StainAug \cite{21_tellez2019cpath} methods consistently outperform all other methods on average, both in terms of F1 score (87.7\%, 86.5\%) and accuracy (88.9\%, 87.4\%). This advantage is particularly pronounced in the MIDOG22  and HISTOPANTUM datasets. The third-ranking algorithm, ARM \cite{zhang2021arm}, has done relatively worse on the MIDOG22  dataset. Furthermore, the Macenko stain normalization algorithm ranked 6\textsuperscript{th}, outperforming 24 other DG algorithms, but not as good as StainAug.

All algorithms performed exceptionally well on the CAMELYON17 dataset (F1$>$90\%), which can be attributed to the abundance of data to help the model generalize better and the relatively simpler nature of the problem. The high performance across algorithms indicates that the CAMELYON17 dataset poses fewer challenges in terms of DS. The DS in CAMELYON17 is primarily stain variation between different hospitals, making StainAug one of the best candidates to improve DG in this dataset (as \cref{tab:full_res} shows the highest F1 96.1\% for StainAug).

The tasks associated with the MIDOG22 and HISTOPANTUM datasets are more challenging, involving more significant domain shifts.
Specifically, the performance metrics in the MIDOG22 task are generally lower compared to other tasks, reflecting the increased difficulty.

Notably, the baseline algorithm, ERM\cite{vapnik1999erm},  demonstrated strong performance (ranked 17\textsuperscript{th}), comparable to other SOTA methods.
This suggests that combining simple augmentations with the ERM approach is sufficient to train a robust classifier.
On the other hand, SANDMask \cite{shahtalebi2021sand} and IGA \cite{koyama2020iga} algorithms struggled to converge on the CAMELYON17 and all datasets, respectively, indicating potential issues in handling domain shifts or complexities in these tasks.


\begin{table*}[th]
\centering
\caption{Benchmarking results for full-scale datasets. In each column, cells are colored from red to green representing worst to best performance.}
\label{tab:full_res}
\begin{tabular}{lcc|cc|cc|cc}
\hline
                            & \multicolumn{2}{c}{CAMELYON17}                                        & \multicolumn{2}{c}{MIDOG22 }                                          & \multicolumn{2}{c}{HISTOPANTUM}                                     & \multicolumn{2}{c}{Average}                                     \\ \cline{2-9} 
\multirow{-2}{*}{Algorithm} & \multicolumn{1}{c}{ACC}          & F1                               & ACC                              & F1                                & ACC                              & F1                               & ACC                          & F1                           \\ \hline
SSL                         & \cellcolor[HTML]{A9D27F}95.4±0.2 & \cellcolor[HTML]{A3D17F}95.2±0.2  & \cellcolor[HTML]{86C87D}79.9±0.2 & \cellcolor[HTML]{63BE7B}76.1±0.6 & \cellcolor[HTML]{63BE7B}91.2±0.6 & \cellcolor[HTML]{63BE7B}91.7±0.5 & \cellcolor[HTML]{63BE7B}88.9 & \cellcolor[HTML]{63BE7B}87.7 \\
StainAug                    & \cellcolor[HTML]{63BE7B}96.4±0.9 & \cellcolor[HTML]{63BE7B}96.1±0.9  & \cellcolor[HTML]{86C87D}79.9±0.3 & \cellcolor[HTML]{6CC17C}76.0±0.4 & \cellcolor[HTML]{A3D17F}85.9±0.0 & \cellcolor[HTML]{9FD07F}87.3±0.0 & \cellcolor[HTML]{93CC7E}87.4 & \cellcolor[HTML]{8ACA7E}86.5 \\
ARM                         & \cellcolor[HTML]{D9E182}94.7±0.3 & \cellcolor[HTML]{D5DF82}94.5±0.3  & \cellcolor[HTML]{FEE983}78.5±0.3 & \cellcolor[HTML]{FEE182}73.4±0.2 & \cellcolor[HTML]{8ECB7E}87.6±0.8 & \cellcolor[HTML]{8ECB7E}88.6±0.8 & \cellcolor[HTML]{A0D07F}87.0 & \cellcolor[HTML]{AAD380}85.5 \\
CausIRL\_CORAL              & \cellcolor[HTML]{FEE883}93.3±0.6 & \cellcolor[HTML]{FEE883}92.8±0.6  & \cellcolor[HTML]{FAEA84}78.9±0.3 & \cellcolor[HTML]{CEDD82}74.9±0.5 & \cellcolor[HTML]{A9D27F}85.4±0.2 & \cellcolor[HTML]{A2D17F}87.1±0.2 & \cellcolor[HTML]{C3DA81}85.9 & \cellcolor[HTML]{BED881}84.9 \\
SelfReg                     & \cellcolor[HTML]{E0E383}94.6±0.3 & \cellcolor[HTML]{E3E383}94.3±0.4  & \cellcolor[HTML]{D7E082}79.2±0.3 & \cellcolor[HTML]{E9E583}74.6±0.4 & \cellcolor[HTML]{C5DB81}83.0±0.4 & \cellcolor[HTML]{BCD881}85.2±0.4 & \cellcolor[HTML]{CDDD82}85.6 & \cellcolor[HTML]{C4DA81}84.7 \\
Macenko                     & \cellcolor[HTML]{FEE883}93.3±0.3 & \cellcolor[HTML]{FEE883}92.9±0.2  & \cellcolor[HTML]{FEE683}77.8±0.3 & \cellcolor[HTML]{FEE683}73.9±0.1 & \cellcolor[HTML]{B4D680}84.4±0.0 & \cellcolor[HTML]{AED480}86.2±0.0 & \cellcolor[HTML]{D9E082}85.2 & \cellcolor[HTML]{D1DE82}84.3 \\
Transfer                    & \cellcolor[HTML]{FEE983}93.7±0.4 & \cellcolor[HTML]{FEEA83}93.6±0.7  & \cellcolor[HTML]{FEE983}78.5±0.3 & \cellcolor[HTML]{FEE983}74.2±0.6 & \cellcolor[HTML]{CADC81}82.6±1.2 & \cellcolor[HTML]{CDDD82}83.9±0.8 & \cellcolor[HTML]{E3E383}84.9 & \cellcolor[HTML]{DEE283}83.9 \\
TRM                         & \cellcolor[HTML]{FEE983}93.5±0.3 & \cellcolor[HTML]{FEE983}93.2±0.4  & \cellcolor[HTML]{C0D981}79.4±0.4 & \cellcolor[HTML]{FBEA84}74.4±0.3 & \cellcolor[HTML]{BFD981}83.5±1.6 & \cellcolor[HTML]{C9DC81}84.2±1.9 & \cellcolor[HTML]{D0DE82}85.5 & \cellcolor[HTML]{DEE283}83.9 \\
IB\_ERM                     & \cellcolor[HTML]{FEEA83}94.0±0.1 & \cellcolor[HTML]{F8E984}94.0±0.2  & \cellcolor[HTML]{B4D680}79.5±0.2 & \cellcolor[HTML]{FEE983}74.2±0.3 & \cellcolor[HTML]{D8E082}81.4±0.6 & \cellcolor[HTML]{DCE182}82.8±0.5 & \cellcolor[HTML]{E0E283}85.0 & \cellcolor[HTML]{E4E483}83.7 \\
CondCAD                     & \cellcolor[HTML]{FEE983}93.5±0.1 & \cellcolor[HTML]{FEE983}93.3±0.1  & \cellcolor[HTML]{FEEA83}78.7±0.3 & \cellcolor[HTML]{E9E583}74.6±0.6 & \cellcolor[HTML]{DEE283}80.9±0.7 & \cellcolor[HTML]{DEE283}82.7±1.2 & \cellcolor[HTML]{F6E984}84.3 & \cellcolor[HTML]{E7E583}83.6 \\
ANDMask                     & \cellcolor[HTML]{FEE983}93.7±0.2 & \cellcolor[HTML]{FEEA83}93.5±0.3  & \cellcolor[HTML]{FEE983}78.6±0.2 & \cellcolor[HTML]{F2E884}74.5±0.3 & \cellcolor[HTML]{FFEB84}78.1±1.0 & \cellcolor[HTML]{EAE583}81.8±0.6 & \cellcolor[HTML]{FEE883}83.5 & \cellcolor[HTML]{F4E884}83.2 \\
Mixup                       & \cellcolor[HTML]{D2DF82}94.8±0.2 & \cellcolor[HTML]{D5DF82}94.5±0.2  & \cellcolor[HTML]{C0D981}79.4±0.3 & \cellcolor[HTML]{E0E283}74.7±0.3 & \cellcolor[HTML]{FBEA84}78.5±0.8 & \cellcolor[HTML]{FDEB84}80.4±0.0 & \cellcolor[HTML]{F9EA84}84.2 & \cellcolor[HTML]{F4E884}83.2 \\
EQRM                        & \cellcolor[HTML]{B0D580}95.3±0.1 & \cellcolor[HTML]{AAD380}95.1±0.1  & \cellcolor[HTML]{63BE7B}80.2±0.1 & \cellcolor[HTML]{B4D680}75.2±0.1 & \cellcolor[HTML]{FEE783}77.4±0.2 & \cellcolor[HTML]{FEDE81}78.9±0.4 & \cellcolor[HTML]{F6E984}84.3 & \cellcolor[HTML]{F7E984}83.1 \\
CausIRL\_MMD                & \cellcolor[HTML]{EEE683}94.4±0.1 & \cellcolor[HTML]{EAE583}94.2±0.1  & \cellcolor[HTML]{FEEA83}78.7±0.6 & \cellcolor[HTML]{FEDC81}72.9±1.9 & \cellcolor[HTML]{F5E884}79.0±2.3 & \cellcolor[HTML]{EAE583}81.8±1.8 & \cellcolor[HTML]{FFEB84}84.0 & \cellcolor[HTML]{FBEA84}83.0 \\
CORAL                       & \cellcolor[HTML]{7FC77D}96.0±0.3 & \cellcolor[HTML]{72C37C}95.9±0.3  & \cellcolor[HTML]{D7E082}79.2±0.4 & \cellcolor[HTML]{CEDD82}74.9±0.2 & \cellcolor[HTML]{FEE583}77.1±0.3 & \cellcolor[HTML]{FDD57F}78.0±0.9 & \cellcolor[HTML]{FCEB84}84.1 & \cellcolor[HTML]{FEEB84}82.9 \\
VREx                        & \cellcolor[HTML]{FCEB84}94.2±0.6 & \cellcolor[HTML]{FEEA83}93.8±0.7  & \cellcolor[HTML]{CCDD82}79.3±0.2 & \cellcolor[HTML]{FEEA83}74.3±0.2 & \cellcolor[HTML]{ECE683}79.7±0.7 & \cellcolor[HTML]{FDEB84}80.4±0.8 & \cellcolor[HTML]{F3E884}84.4 & \cellcolor[HTML]{FEEA83}82.8 \\
\hdashline
ERM                         & \cellcolor[HTML]{9BCF7F}95.6±0.0 & \cellcolor[HTML]{95CD7E}95.4±0.0  & \cellcolor[HTML]{E3E383}79.1±0.2 & \cellcolor[HTML]{E0E283}74.7±0.2 & \cellcolor[HTML]{FEE482}76.8±0.1 & \cellcolor[HTML]{FDD17F}77.6±0.9 & \cellcolor[HTML]{FEEA83}83.8 & \cellcolor[HTML]{FEE983}82.6 \\
\hdashline
GroupDRO                    & \cellcolor[HTML]{E0E383}94.6±0.6 & \cellcolor[HTML]{DCE182}94.4±0.8  & \cellcolor[HTML]{FEEA83}78.8±0.1 & \cellcolor[HTML]{F2E884}74.5±0.4 & \cellcolor[HTML]{FEE983}77.8±0.3 & \cellcolor[HTML]{FEDB80}78.6±0.1 & \cellcolor[HTML]{FEE983}83.7 & \cellcolor[HTML]{FEE883}82.5 \\
CAD                         & \cellcolor[HTML]{FEE983}93.7±0.2 & \cellcolor[HTML]{FEE983}93.4±0.3  & \cellcolor[HTML]{FEE582}77.6±0.2 & \cellcolor[HTML]{FEE282}73.5±0.3 & \cellcolor[HTML]{FEE883}77.7±3.1 & \cellcolor[HTML]{FFEB84}80.2±2.8 & \cellcolor[HTML]{FEE683}83.0 & \cellcolor[HTML]{FEE783}82.4 \\
MTL                         & \cellcolor[HTML]{FEEA83}93.8±0.0 & \cellcolor[HTML]{FEE983}93.4±0.0  & \cellcolor[HTML]{FAEA84}78.9±0.4 & \cellcolor[HTML]{FEE683}73.9±0.7 & \cellcolor[HTML]{FFEB84}78.1±1.7 & \cellcolor[HTML]{FEE883}79.9±2.3 & \cellcolor[HTML]{FEE983}83.6 & \cellcolor[HTML]{FEE783}82.4 \\
MLDG                        & \cellcolor[HTML]{B7D780}95.2±0.3 & \cellcolor[HTML]{B9D780}94.9±0.4  & \cellcolor[HTML]{FAEA84}78.9±0.3 & \cellcolor[HTML]{FEE983}74.2±0.4 & \cellcolor[HTML]{FEDD81}75.5±0.2 & \cellcolor[HTML]{FDD17F}77.6±0.4 & \cellcolor[HTML]{FEE783}83.2 & \cellcolor[HTML]{FEE783}82.3 \\
RSC                         & \cellcolor[HTML]{FEEA83}94.1±0.1 & \cellcolor[HTML]{FEEA83}93.7±0.2  & \cellcolor[HTML]{FEEA83}78.8±0.4 & \cellcolor[HTML]{87C97E}75.7±0.2 & \cellcolor[HTML]{FEE382}76.6±1.2 & \cellcolor[HTML]{FDCE7E}77.3±0.4 & \cellcolor[HTML]{FEE783}83.2 & \cellcolor[HTML]{FEE683}82.2 \\
SD                          & \cellcolor[HTML]{A2D17F}95.5±0.3 & \cellcolor[HTML]{9CCF7F}95.3±0.4  & \cellcolor[HTML]{FEE983}78.5±0.6 & \cellcolor[HTML]{FEE683}73.9±0.3 & \cellcolor[HTML]{FFEB84}78.1±0.9 & \cellcolor[HTML]{FDCE7E}77.3±0.9 & \cellcolor[HTML]{FFEB84}84.0 & \cellcolor[HTML]{FEE683}82.2 \\
IRM                         & \cellcolor[HTML]{CCDD82}94.9±0.4 & \cellcolor[HTML]{C7DB81}94.7±0.5  & \cellcolor[HTML]{FEE783}78.1±0.5 & \cellcolor[HTML]{FEDC81}72.9±0.5 & \cellcolor[HTML]{FEE883}77.6±0.4 & \cellcolor[HTML]{FEDE81}78.9±0.3 & \cellcolor[HTML]{FEE883}83.5 & \cellcolor[HTML]{FEE583}82.1 \\
CDANN                       & \cellcolor[HTML]{FEE482}91.6±1.2 & \cellcolor[HTML]{FEE482}90.9±1.4  & \cellcolor[HTML]{FEEA83}78.8±0.6 & \cellcolor[HTML]{FEEA83}74.3±0.3 & \cellcolor[HTML]{E8E583}80.1±1.9 & \cellcolor[HTML]{FFEB84}80.2±2.1 & \cellcolor[HTML]{FEE883}83.5 & \cellcolor[HTML]{FEE382}81.8 \\
SagNet                      & \cellcolor[HTML]{FEE983}93.6±0.0 & \cellcolor[HTML]{FEE983}93.2±0.0  & \cellcolor[HTML]{FEEA83}78.7±0.2 & \cellcolor[HTML]{D7E082}74.8±0.3 & \cellcolor[HTML]{FEE182}76.2±1.0 & \cellcolor[HTML]{FDCB7D}77.0±0.9 & \cellcolor[HTML]{FEE683}82.8 & \cellcolor[HTML]{FEE282}81.6 \\
DANN                        & \cellcolor[HTML]{FEE482}91.8±1.8 & \cellcolor[HTML]{FEE582}91.5±1.8  & \cellcolor[HTML]{D7E082}79.2±0.3 & \cellcolor[HTML]{FEE983}74.2±0.4 & \cellcolor[HTML]{FFEB84}78.1±0.5 & \cellcolor[HTML]{FDCB7D}77.0±0.8 & \cellcolor[HTML]{FEE683}83.0 & \cellcolor[HTML]{FEDD81}80.9 \\
MMD                         & \cellcolor[HTML]{FCEB84}94.2±0.4 & \cellcolor[HTML]{E3E383}94.3±0.2  & \cellcolor[HTML]{FEDB80}75.3±1.8 & \cellcolor[HTML]{FCB77A}69.0±2.7 & \cellcolor[HTML]{FEE983}77.8±0.1 & \cellcolor[HTML]{FEDD81}78.8±0.1 & \cellcolor[HTML]{FEE482}82.4 & \cellcolor[HTML]{FEDC81}80.7 \\
IGA                         & \cellcolor[HTML]{F98570}55.5±0.9 & \cellcolor[HTML]{FBAF78}67.7±0.3  & \cellcolor[HTML]{F8696B}49.4±3.5 & \cellcolor[HTML]{F8696B}60.9±0.0 & \cellcolor[HTML]{F8696B}52.8±2.5 & \cellcolor[HTML]{F8696B}67.1±0.9 & \cellcolor[HTML]{F8696B}52.6 & \cellcolor[HTML]{F8736D}65.2 \\
SANDMask                    & \cellcolor[HTML]{F8696B}44.6±4.9 & \cellcolor[HTML]{F8696B}37.1±13.7 & \cellcolor[HTML]{C0D981}79.4±0.6 & \cellcolor[HTML]{E0E283}74.7±0.4 & \cellcolor[HTML]{FEE983}77.9±0.3 & \cellcolor[HTML]{FEDF81}79.0±0.9 & \cellcolor[HTML]{FBA576}67.3 & \cellcolor[HTML]{F8696B}63.6 \\ \hline
\end{tabular}
\end{table*}

\begin{table*}[t]
\centering
\caption{Benchmarking results for small-scale datasets. In each column, cells are colored from red to green representing worst to best performance.}
\label{tab:small_res}
\begin{tabular}{lcc|cc|cc|cc}
\hline
                            & \multicolumn{2}{c}{sCAMELYON17}                                        & \multicolumn{2}{c}{sMIDOG22}                                          & \multicolumn{2}{c}{sHISTOPANTUM}                                     & \multicolumn{2}{c}{Average}                                     \\ \cline{2-9} 
\multirow{-2}{*}{Algorithm} & \multicolumn{1}{c}{ACC}          & F1                               & ACC                              & F1                                & ACC                              & F1                               & ACC                          & F1                           \\ \hline
SSL                         & \cellcolor[HTML]{63BE7B}94.9±0.3 & \cellcolor[HTML]{63BE7B}93.2±1.0 & \cellcolor[HTML]{FEEA83}76.6±0.7 & \cellcolor[HTML]{FFEB84}71.1±0.7  & \cellcolor[HTML]{63BE7B}91.6±0.5 & \cellcolor[HTML]{63BE7B}92.0±0.5 & \cellcolor[HTML]{63BE7B}87.7 & \cellcolor[HTML]{63BE7B}85.4 \\
Transfer                    & \cellcolor[HTML]{CFDE82}90.1±0.8 & \cellcolor[HTML]{C9DC81}88.7±0.7 & \cellcolor[HTML]{63BE7B}77.9±0.3 & \cellcolor[HTML]{63BE7B}73.7±0.6  & \cellcolor[HTML]{FEE983}85.1±0.2 & \cellcolor[HTML]{FEE983}86.0±0.1 & \cellcolor[HTML]{DDE283}84.3 & \cellcolor[HTML]{BFD981}82.8 \\
StainAug                    & \cellcolor[HTML]{CDDD82}90.2±0.0 & \cellcolor[HTML]{D0DE82}88.4±0.8 & \cellcolor[HTML]{DEE283}77.0±0.2 & \cellcolor[HTML]{BDD881}72.2±1.0  & \cellcolor[HTML]{E7E583}86.4±0.1 & \cellcolor[HTML]{DEE283}87.4±0.1 & \cellcolor[HTML]{D6E082}84.5 & \cellcolor[HTML]{C2DA81}82.7 \\
Macenko                     & \cellcolor[HTML]{A5D17F}92.0±0.5 & \cellcolor[HTML]{A5D17F}90.3±0.7 & \cellcolor[HTML]{FEE683}75.7±0.7 & \cellcolor[HTML]{FEE883}70.5±0.7  & \cellcolor[HTML]{FCEA84}85.6±0.2 & \cellcolor[HTML]{EBE683}86.9±0.1 & \cellcolor[HTML]{D6E082}84.5 & \cellcolor[HTML]{C6DB81}82.6 \\
VREx                        & \cellcolor[HTML]{CDDD82}90.2±0.7 & \cellcolor[HTML]{D0DE82}88.4±1.4 & \cellcolor[HTML]{FEE683}75.7±0.2 & \cellcolor[HTML]{D5DF82}71.8±0.3  & \cellcolor[HTML]{E2E383}86.6±1.1 & \cellcolor[HTML]{E1E383}87.3±1.0 & \cellcolor[HTML]{E1E383}84.2 & \cellcolor[HTML]{C9DC81}82.5 \\
SagNet                      & \cellcolor[HTML]{CFDE82}90.1±0.3 & \cellcolor[HTML]{C5DB81}88.9±0.2 & \cellcolor[HTML]{FEE582}75.6±0.7 & \cellcolor[HTML]{BDD881}72.2±0.5  & \cellcolor[HTML]{FEE783}84.7±0.4 & \cellcolor[HTML]{FEE683}85.5±0.3 & \cellcolor[HTML]{FAEA84}83.5 & \cellcolor[HTML]{D4DF82}82.2 \\
IB\_ERM                     & \cellcolor[HTML]{FEEA83}87.9±1.0 & \cellcolor[HTML]{D2DE82}88.3±0.6 & \cellcolor[HTML]{EBE683}76.9±0.5 & \cellcolor[HTML]{FFEB84}71.1±1.7  & \cellcolor[HTML]{ECE683}86.2±0.4 & \cellcolor[HTML]{E6E483}87.1±0.3 & \cellcolor[HTML]{F3E884}83.7 & \cellcolor[HTML]{D7E082}82.1 \\
\hdashline
ERM                         & \cellcolor[HTML]{FAEA84}88.2±0.4 & \cellcolor[HTML]{FEE883}85.8±0.4 & \cellcolor[HTML]{B5D680}77.3±0.3 & \cellcolor[HTML]{A6D27F}72.6±0.5  & \cellcolor[HTML]{D6DF82}87.1±0.1 & \cellcolor[HTML]{E1E383}87.3±0.3 & \cellcolor[HTML]{E1E383}84.2 & \cellcolor[HTML]{DEE283}81.9 \\
\hdashline
GroupDRO                    & \cellcolor[HTML]{FCEB84}88.1±1.6 & \cellcolor[HTML]{E0E283}87.7±1.0 & \cellcolor[HTML]{DEE283}77.0±0.2 & \cellcolor[HTML]{FFEB84}71.1±0.1  & \cellcolor[HTML]{F7E984}85.8±0.2 & \cellcolor[HTML]{FEE983}86.0±0.6 & \cellcolor[HTML]{F3E884}83.7 & \cellcolor[HTML]{E9E583}81.6 \\
ANDMask                     & \cellcolor[HTML]{BDD881}90.9±0.6 & \cellcolor[HTML]{BED981}89.2±0.6 & \cellcolor[HTML]{FEE683}75.8±0.1 & \cellcolor[HTML]{E7E583}71.5±1.0  & \cellcolor[HTML]{FEDD81}82.6±0.2 & \cellcolor[HTML]{FED980}83.7±0.0 & \cellcolor[HTML]{FEE983}83.1 & \cellcolor[HTML]{ECE683}81.5 \\
TRM                         & \cellcolor[HTML]{FEEA83}87.9±0.7 & \cellcolor[HTML]{FEE983}86.0±0.7 & \cellcolor[HTML]{8CCA7E}77.6±0.2 & \cellcolor[HTML]{FFEB84}71.1±0.6  & \cellcolor[HTML]{DDE283}86.8±0.3 & \cellcolor[HTML]{E4E383}87.2±0.1 & \cellcolor[HTML]{E5E483}84.1 & \cellcolor[HTML]{F0E784}81.4 \\
CDANN                       & \cellcolor[HTML]{EAE583}88.9±0.2 & \cellcolor[HTML]{D9E082}88.0±0.2 & \cellcolor[HTML]{FEE783}76.0±0.8 & \cellcolor[HTML]{FEE683}70.1±0.0  & \cellcolor[HTML]{FEE683}84.5±0.2 & \cellcolor[HTML]{FEE783}85.7±0.9 & \cellcolor[HTML]{FEE983}83.1 & \cellcolor[HTML]{F3E884}81.3 \\
ARM                         & \cellcolor[HTML]{F8E984}88.3±0.8 & \cellcolor[HTML]{FEE783}85.7±0.7 & \cellcolor[HTML]{D0DE82}77.1±1.0 & \cellcolor[HTML]{F3E884}71.3±0.6  & \cellcolor[HTML]{EFE784}86.1±0.6 & \cellcolor[HTML]{FCEA84}86.3±1.0 & \cellcolor[HTML]{EFE784}83.8 & \cellcolor[HTML]{FAEA84}81.1 \\
EQRM                        & \cellcolor[HTML]{FEE883}87.3±0.1 & \cellcolor[HTML]{FEE382}85.0±0.2 & \cellcolor[HTML]{D0DE82}77.1±0.2 & \cellcolor[HTML]{DBE182}71.7±0.2  & \cellcolor[HTML]{FEE883}85.0±1.1 & \cellcolor[HTML]{FEEB84}86.2±0.7 & \cellcolor[HTML]{FEE983}83.1 & \cellcolor[HTML]{FEEB84}81.0 \\
SelfReg                     & \cellcolor[HTML]{FEE883}87.2±1.7 & \cellcolor[HTML]{FEDF81}84.4±2.8 & \cellcolor[HTML]{F9EA84}76.8±0.2 & \cellcolor[HTML]{E2E383}71.6±0.1  & \cellcolor[HTML]{EFE784}86.1±0.4 & \cellcolor[HTML]{E6E483}87.1±0.5 & \cellcolor[HTML]{FEEB84}83.4 & \cellcolor[HTML]{FEEB84}81.0 \\
MTL                         & \cellcolor[HTML]{E1E383}89.3±1.1 & \cellcolor[HTML]{E4E483}87.5±1.6 & \cellcolor[HTML]{8CCA7E}77.6±0.1 & \cellcolor[HTML]{EDE683}71.4±1.1  & \cellcolor[HTML]{FEE282}83.5±1.1 & \cellcolor[HTML]{FEDA80}83.9±0.5 & \cellcolor[HTML]{FAEA84}83.5 & \cellcolor[HTML]{FEEA83}80.9 \\
MLDG                        & \cellcolor[HTML]{FEE883}87.4±0.6 & \cellcolor[HTML]{F9EA84}86.6±0.5 & \cellcolor[HTML]{FEE983}76.4±0.3 & \cellcolor[HTML]{E7E583}71.5±1.4  & \cellcolor[HTML]{FEDD81}82.5±1.5 & \cellcolor[HTML]{FEDD81}84.2±0.7 & \cellcolor[HTML]{FEE582}82.1 & \cellcolor[HTML]{FEE983}80.8 \\
CausIRL\_CORAL              & \cellcolor[HTML]{FEE683}86.8±0.9 & \cellcolor[HTML]{FEE282}84.8±1.5 & \cellcolor[HTML]{F9EA84}76.8±0.2 & \cellcolor[HTML]{FEE482}69.8±0.6  & \cellcolor[HTML]{E7E583}86.4±0.8 & \cellcolor[HTML]{E4E383}87.2±0.8 & \cellcolor[HTML]{FEEA83}83.3 & \cellcolor[HTML]{FEE883}80.6 \\
SANDMask                    & \cellcolor[HTML]{FEEB84}88.0±0.6 & \cellcolor[HTML]{F4E884}86.8±0.1 & \cellcolor[HTML]{FEE783}76.0±0.3 & \cellcolor[HTML]{FEE983}70.8±1.2  & \cellcolor[HTML]{FEE182}83.3±0.2 & \cellcolor[HTML]{FEDC81}84.1±0.5 & \cellcolor[HTML]{FEE683}82.4 & \cellcolor[HTML]{FEE883}80.6 \\
SD                          & \cellcolor[HTML]{FEE382}86.0±1.1 & \cellcolor[HTML]{FDD780}82.9±1.5 & \cellcolor[HTML]{F9EA84}76.8±0.6 & \cellcolor[HTML]{9ACE7F}72.8±0.0  & \cellcolor[HTML]{FEEA83}85.4±0.5 & \cellcolor[HTML]{FEEB84}86.2±0.3 & \cellcolor[HTML]{FEE883}82.7 & \cellcolor[HTML]{FEE883}80.6 \\
CondCAD                     & \cellcolor[HTML]{FEE783}87.0±0.0 & \cellcolor[HTML]{FEE282}84.8±0.1 & \cellcolor[HTML]{9ACE7F}77.5±0.2 & \cellcolor[HTML]{FEE683}70.1±0.7  & \cellcolor[HTML]{F9EA84}85.7±1.6 & \cellcolor[HTML]{F4E884}86.6±1.4 & \cellcolor[HTML]{FEEB84}83.4 & \cellcolor[HTML]{FEE783}80.5 \\
Mixup                       & \cellcolor[HTML]{FEE683}86.8±0.6 & \cellcolor[HTML]{FEE081}84.5±1.1 & \cellcolor[HTML]{FEEA83}76.7±0.4 & \cellcolor[HTML]{FEE983}70.7±1.4  & \cellcolor[HTML]{FEE983}85.1±0.2 & \cellcolor[HTML]{FEE983}86.0±0.0 & \cellcolor[HTML]{FEE983}82.9 & \cellcolor[HTML]{FEE683}80.4 \\
RSC                         & \cellcolor[HTML]{FEE783}87.0±2.0 & \cellcolor[HTML]{FEE182}84.6±2.8 & \cellcolor[HTML]{C2DA81}77.2±0.6 & \cellcolor[HTML]{FEE683}70.2±0.4  & \cellcolor[HTML]{F4E884}85.9±1.9 & \cellcolor[HTML]{FEEA83}86.1±0.9 & \cellcolor[HTML]{FEEB84}83.4 & \cellcolor[HTML]{FEE583}80.3 \\
CAD                         & \cellcolor[HTML]{FEEA83}87.7±2.0 & \cellcolor[HTML]{FEE783}85.7±2.8 & \cellcolor[HTML]{FEE082}74.6±0.4 & \cellcolor[HTML]{FEE482}69.8±0.1  & \cellcolor[HTML]{FEE582}84.3±0.4 & \cellcolor[HTML]{FEE282}84.9±0.1 & \cellcolor[HTML]{FEE683}82.2 & \cellcolor[HTML]{FEE482}80.1 \\
CORAL                       & \cellcolor[HTML]{FEE182}85.3±0.5 & \cellcolor[HTML]{FED980}83.2±1.8 & \cellcolor[HTML]{9ACE7F}77.5±0.6 & \cellcolor[HTML]{FEE582}69.9±0.1  & \cellcolor[HTML]{F2E784}86.0±0.1 & \cellcolor[HTML]{FEE983}85.9±0.3 & \cellcolor[HTML]{FEE983}82.9 & \cellcolor[HTML]{FEE182}79.7 \\
IRM                         & \cellcolor[HTML]{FCEB84}88.1±0.4 & \cellcolor[HTML]{F4E884}86.8±1.2 & \cellcolor[HTML]{FEE783}76.0±0.2 & \cellcolor[HTML]{FDD37F}66.0±1.7  & \cellcolor[HTML]{FEEA83}85.3±1.9 & \cellcolor[HTML]{FEE883}85.8±1.9 & \cellcolor[HTML]{FEE983}83.1 & \cellcolor[HTML]{FEDF81}79.5 \\
DANN                        & \cellcolor[HTML]{FEDF81}84.9±1.7 & \cellcolor[HTML]{FCC47C}79.7±1.3 & \cellcolor[HTML]{FEE082}74.6±0.9 & \cellcolor[HTML]{FEDB80}67.7±1.8  & \cellcolor[HTML]{FEE082}83.2±0.6 & \cellcolor[HTML]{FEDE81}84.4±0.4 & \cellcolor[HTML]{FEE081}80.9 & \cellcolor[HTML]{FDCE7E}77.3 \\
CausIRL\_MMD                & \cellcolor[HTML]{FEE583}86.6±0.7 & \cellcolor[HTML]{FEDC81}83.8±1.5 & \cellcolor[HTML]{FBB078}64.3±8.8 & \cellcolor[HTML]{FCB379}59.2±7.7  & \cellcolor[HTML]{FEEB84}85.5±0.9 & \cellcolor[HTML]{FCEA84}86.3±1.2 & \cellcolor[HTML]{FDD780}78.8 & \cellcolor[HTML]{FDC77D}76.4 \\
MMD                         & \cellcolor[HTML]{D8E082}89.7±0.7 & \cellcolor[HTML]{CCDD82}88.6±1.0 & \cellcolor[HTML]{FBA977}62.9±5.6 & \cellcolor[HTML]{F8696B}43.0±14.4 & \cellcolor[HTML]{FEE883}85.0±0.2 & \cellcolor[HTML]{F4E884}86.6±0.7 & \cellcolor[HTML]{FED980}79.2 & \cellcolor[HTML]{FBAA77}72.7 \\
IGA                         & \cellcolor[HTML]{F8696B}52.9±0.3 & \cellcolor[HTML]{F8696B}63.9±1.0 & \cellcolor[HTML]{F8696B}49.3±1.1 & \cellcolor[HTML]{FCBB7A}60.9±0.2  & \cellcolor[HTML]{F8696B}57.1±0.8 & \cellcolor[HTML]{F8696B}67.9±0.6 & \cellcolor[HTML]{F8696B}53.1 & \cellcolor[HTML]{F8696B}64.2 \\ \hline
\end{tabular}
\end{table*}

\subsection{Results for Sub-sampled (small) datasets}

To investigate how DG algorithms perform under a low-data budget scenario, we subsample each dataset at different rates to create smaller datasets (as explained in \cref{sec:sub-dataset}) and repeat the experiments. The results for these experiments are reported in \cref{tab:small_res}.

Interestingly, SSL and StainAug algorithms are still among the top 3 performing algorithms with Transfer algorithm \cite{zhang2021transfer} place on the second rank and achieving F1 of 82.8\% (almost on a par with StainAug, F1=82.7\%). However, SSL considerably outperforms other algorithms by gaining the F1 score of 85.4\%, showing an advantage in small-dataset scenarios as has been shown before for other algorithms based on self-supervised learning \cite{89_koohbanani2021cpath}. The majority of SSL superiority is owed to the performance of sCAMELYON17 and sHISTOPANTUM datasets. However, on the hardest DG task using the sMIDOG22 dataset, Transfer algorithm \cite{zhang2021transfer} gains the highest F1 of 77.9\%, considerably outperforming SSL. StainAug does not perform as high as SSL on sHISTOPANTUM, nevertheless, it keeps a good performance on sMIDOG22 and sCAMELYON. On the sHISTOPANTUM dataset, except for the SSL algorithm, most of the other algorithms perform on par. 

The baseline ERM algorithm archives impressive results in the small-scale dataset, outperforming all other algorithms on the sHISTOPANTUM dataset excluding SSL (F1=87.1\%) and very good performance in the other small datasets.
On the other hand, IGA \cite{koyama2020iga} still struggles to converge on small datasets whereas the SANDMask \cite{shahtalebi2021sand} algorithm works relatively well on sCAMELYON17 dataset although it could not converge on large-scale CAMELYON17 dataset. 

\subsection{Domain-level performance}
\begin{figure*}[!h]
  \centering
  \begin{subfigure}[A]{1\textwidth}
  \caption{Accuracy over different domains}
    \includegraphics[width=\textwidth]{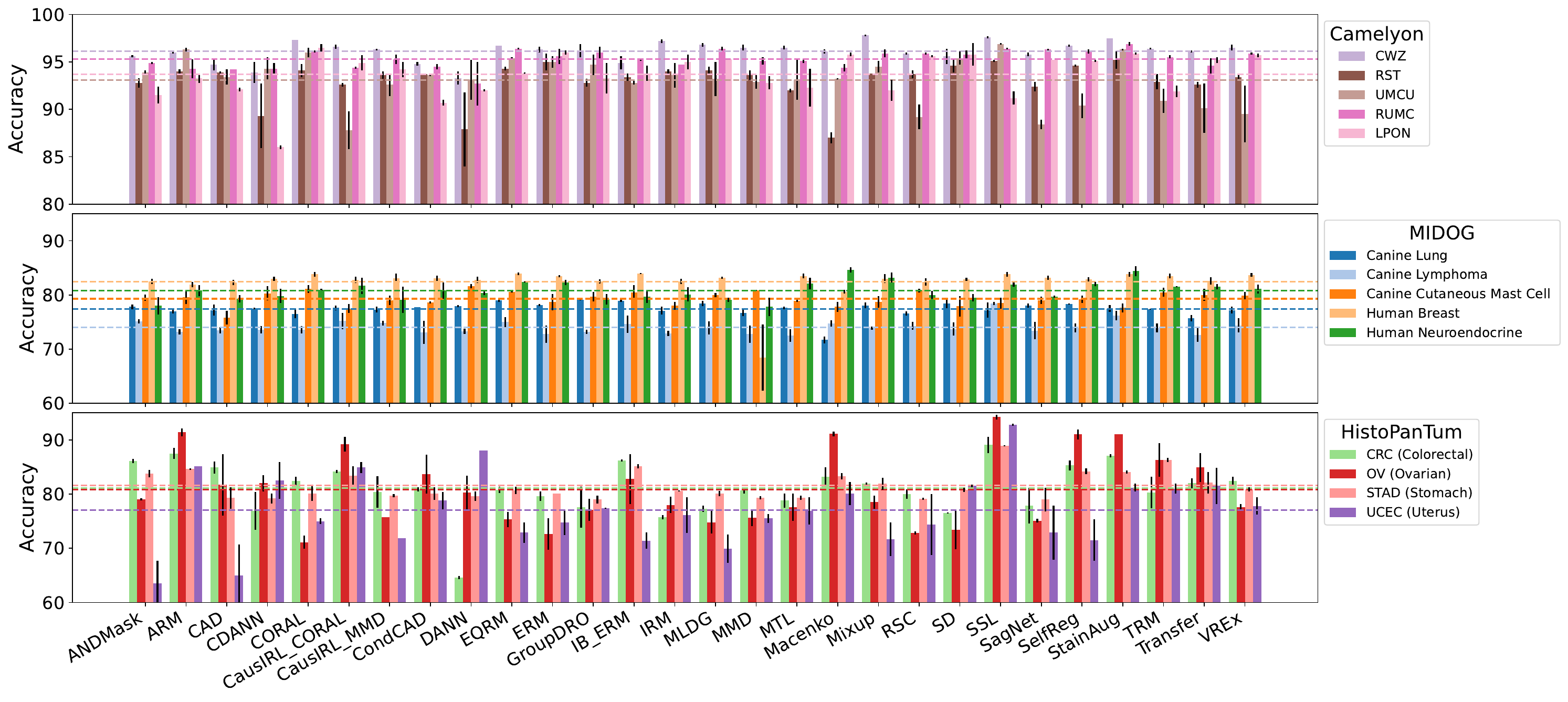}
  \end{subfigure}
  \hfill
  \begin{subfigure}[B]{1\textwidth}
  \caption{F1 Score over different domains}
    \includegraphics[width=\textwidth]
    {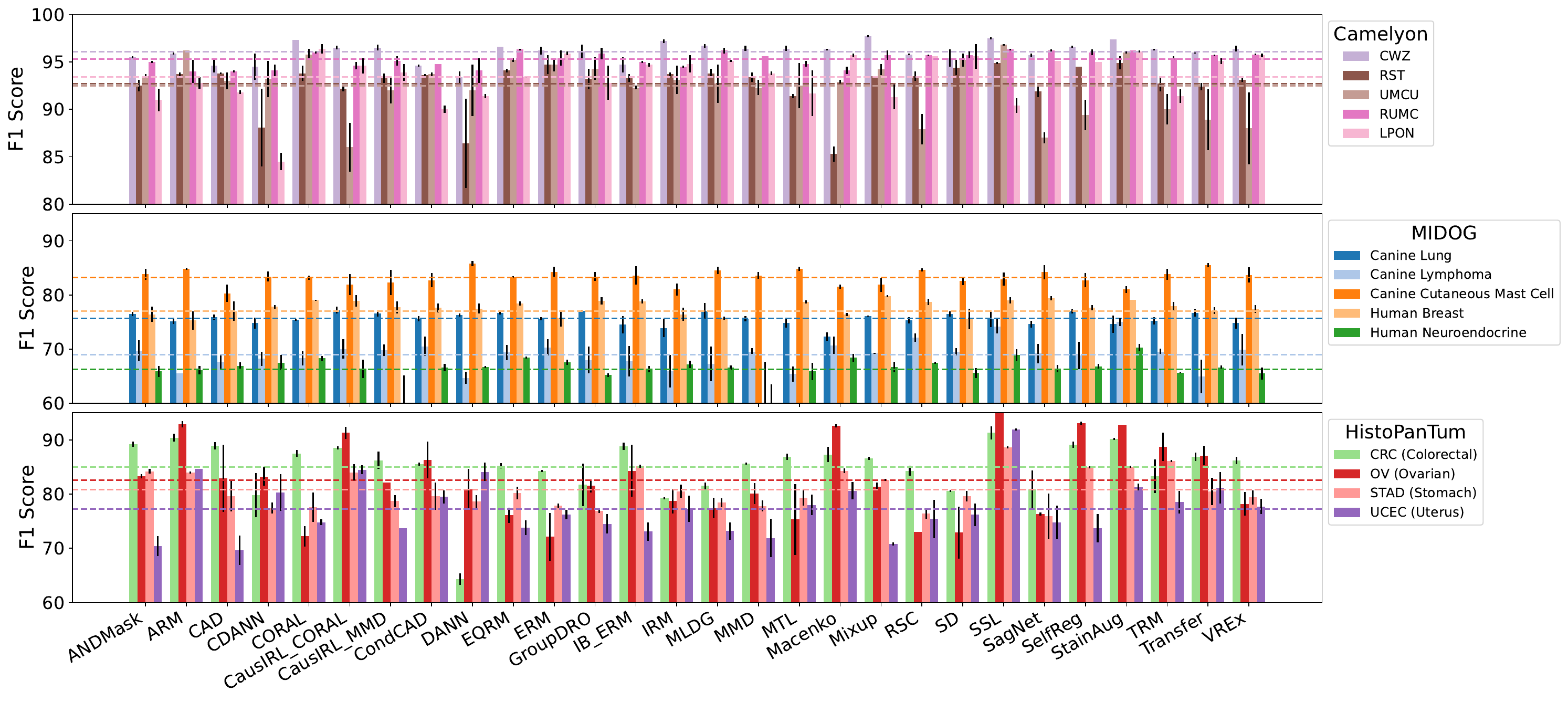}
  \end{subfigure}
  \caption{Benchmarking results for different algorithms, (A) Accuracy and (B) F1 Score. Each domain is presented by a unique color and the average performance of all algorithms over each domain is presented with a horizontal dashed line with the same color.
  }
  \label{fig:res-domains}
\end{figure*}

We comprehensively evaluate the performance of various algorithms across different domains, the results of which are presented in \cref{fig:res-domains} in the form of bar plots. In \cref{fig:res-domains}, each domain in every dataset is represented by a uniquely colored bar.
The average performance of all algorithms over each domain is indicated by horizontal dashed lines in the same color as the domain. In \cref{fig:res-domains}, algorithms "IGA" and "SANDMask" are excluded due to poor performance and to better visualize the working performance range of other algorithms.

\paragraph{CAMELYON17} Performance across centers is generally high, with average F1 scores around 93\% to 96\%. This is highlighted by the closely clustered bars and the horizontal dashed lines at the top of the graph (only a 3.5\% difference between the best and worst domains). Results for CWZ and RUMC centers are consistently among the highest scores, presumably because slides from these two centers were scanned using the same scanner (at RUMC center) and there is a lower domain shift between these two datasets, hence a model trained on the data from one of these domains will perform reasonably good for the other domain too. Furthermore, from \cref{fig:res-domains}B it is evident that StainAug \cite{21_tellez2019cpath}, SD \cite{pezeshki2021sd}, and ERM \cite{vapnik1999erm} are among the most consistent algorithms over different domains whereas the performance of DAN \cite{ganin2016dann}, CDANN \cite{135_li2018general}, CausIRL\_{MMD} \cite{chevalley2022CausIRL}, and Macenko \cite{199_macenko2009cpath} changes considerably for different CAMELYON17 domains.

\paragraph{MIDOG22} In this dataset, performance varies more significantly across the different domains (11\% difference between highest and lowest average performance). In particular, based on the F1 score, the human endocrine cancer is the hardest domain (F1=66\%)  and the canine cutaneous mast cell tumor is the easiest (F1=77\%) for out-of-domain mitosis detection. Interestingly, the patterns of different algorithms' performances over different domains are similar, i.e., best to worst performing domains being canine cutaneous mast cell, human breast,  canine lung, canine lymphoma, and human neuroendocrine. The worst performance on human neuroendocrine can be justified by three reasons: the tumor type is completely different from all other domains (class-conditional shift), slides in this domain were scanned with a Hamamatsu NanoZoomer XR scanner unlike other datasets that used Aperio or 3DHistech scanners (covariate shift), and the label distribution in this domain is significantly different from other domains (prior shift). On the other hand, the performance on the canine cutaneous mast cell domain is higher because in that case 2 other similar tumor types from the same species and scanner are utilized for model training, hence model seeing similar data can generalize better to this unseen domain. In terms of consistency over different domains, although the accuracy metric shows consistency for most algorithms (such as SSL in \cref{fig:res-domains}A), F1 score values tell another story where consistency over different domains drastically decreases for all algorithms. This is mostly due to an imbalanced class population across different domains in the MIDOG22 dataset (see \cref{fig:data}B).

\paragraph{HISTOPANTUM} Average accuracy over different domains in around the same range for CRC, OV, and STAD domains (around 81\%) with a notable accuracy drop for the UCEC domain (77\%). The F1 score on the UCEC domain is also the worst among all domains (77\% compared to 80-85\% for other domains). This can be accounted for by a prior shift in the label distribution when comparing the UCEC domain with others. CRC domain gets the highest performance over because it shares a similar label distribution and tissue phenotypes appearance (especially with the STAD domain). In HISTOPANTUM, the SSL algorithm is the best performing and consistent algorithm across all domains, mostly because the utilized SSL algorithm has already seen TCGA slide during its pertaining procedure. Furthermore, StainAug, ARM \cite{zhang2021arm}, CausIRL\_CORAL \cite{chevalley2022CausIRL} and Transfer \cite{zhang2021transfer} algorithms also show decent consistency and performance over different domains.

\section{Discussion}
\label{sec:disc}

Benchmarking DG algorithms is essential for evaluating their performance across diverse datasets and scenarios to provide insights that can increase model robustness in real-world applications. In this study, we benchmarked various DG algorithms with different working principles, including SOTA algorithms from DomainBed collection \cite{domainbed}, self-supervised learning \cite{159_kang2023cpath}, and pathology-specific techniques \cite{199_macenko2009cpath,21_tellez2019cpath}, on datasets with different DS and size properties. Our unified and fair benchmarking process reported both accuracy and F1 scores for comprehensive evaluation through robust cross-validation experiments.

At a glance, the average best-performing algorithms are SSL and StainAug. SSL methods excelled especially on the HISTOPANTUM dataset, which is the main reason SSL ranked first in the full-scale dataset (\cref{tab:full_res}) and small-scale dataset (\cref{tab:small_res}) scenarios. This is mostly because SSL pertaining was done on an extensive set of patches extracted from TCGA, the same source used to curate the HISTOPANTUM dataset. Although the same image patches and labels are not shared between HISTOPANTUM and the dataset used during pertaining of SSL, the SSL has indirect access to the test data in HISTOPANTUM and already has seen the possible variation of image data within HISTOPANTUM. Therefore, the evaluation of SSL on HISTOPANTUM is more of a ``domain adaptation'' exercise rather than ``domain generalization'' and comparing its performance (F1=91.7\%) with the rest of the algorithms (such as ARM and StainAug with F1 scores of 88.6\% and 87.3\%, respectively) is not fair. Nevertheless, SSL shows excellent performance in CAMELYON17, MIDOG22, and the low-data-budget scenario of the sCAMELYON17 dataset. However, that is not the case with the sMIDOG22 dataset where there are all sorts of DS and data shortage problems.

CPath-specific algorithms--stain augmentation (StainAug) and stain normalization (Macenko)--outperform most complex DG algorithms in the literature while being simple and easy to implement. In particular, StainAug excelled in the CAMELYON17 and MIDOG22 datasets. StainAug helps the model to learn more stain-invariant feature representations from the image during the training by randomly tweaking the stain information on the fly. Considering stain variation as a confounding factor, StainAug can be thought of as a causal approach for DG in CPath as it helps to learn features that are irrelevant to stain variation, hence achieving outstanding results on the CAMELYON17 dataset where the main DS is covariate shift (or changes in stain appearance). Macenko stain normalization algorithm has also shined in the CAMELYON17 and HISTOPANTUM tasks where stain variation is the dominant DS. Specifically, in the small dataset scenario of sCAMELYON, Macenko outperforms all other algorithms (except SSL) and ranks second. However, we should note that stain normalization methods add another preprocessing step to every CPath pipeline that uses them and sometimes are not stable in practice \cite{82_vu2022cpath,jahanifar2023domain}. Therefore, we suggest using stain augmentation over stain normalization when training models on H\&E images. 


\begin{figure*}[h]
  \centering
  \begin{subfigure}[b]{0.4\textwidth}
    \includegraphics[width=\textwidth]{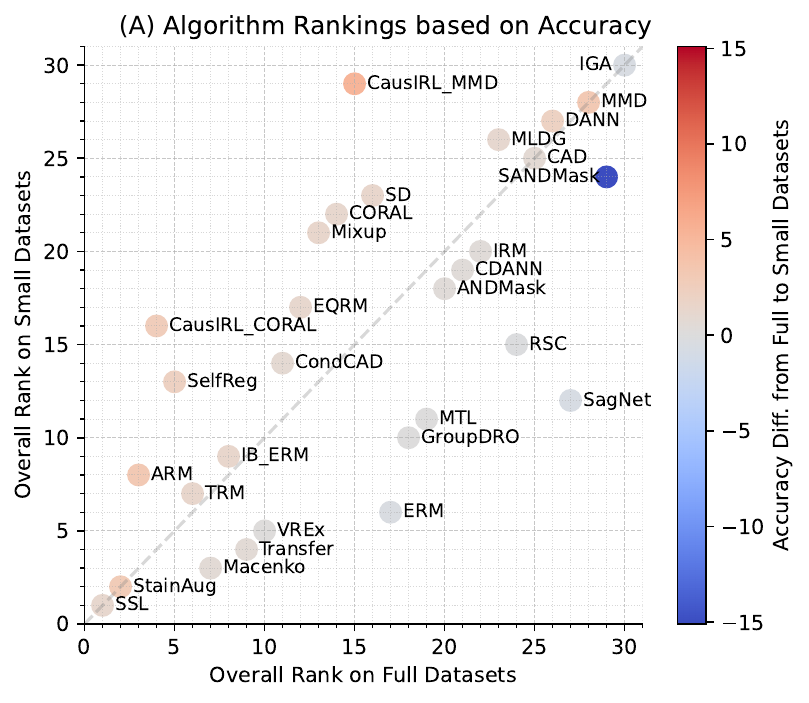}
  \end{subfigure}
  \hfill
  \begin{subfigure}[b]{0.4\textwidth}
    \includegraphics[width=\textwidth]{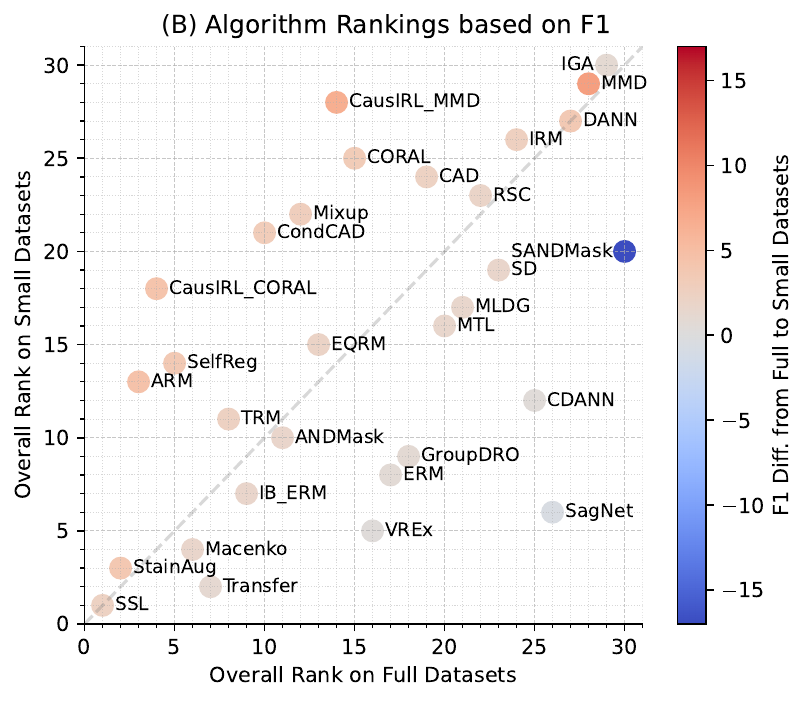}
  \end{subfigure}
  \caption{Comparative analysis of algorithm performance in small vs. full dataset regimes, showcasing Accuracy (A) and F1 score (B) metrics. The plots illustrate how each algorithm's effectiveness varies with dataset size. Algorithms close to the bottom-left corner are desirable.}
  \label{fig:rank-dff}
\end{figure*}

To gain a deeper understanding of how algorithm performance varies between small and full dataset regimes, we plot their rankings based on F1 and Accuracy metrics over small and full datasets in \cref{fig:rank-dff}.  
Each point on the plot represents an algorithm, with its x-axis position indicating its rank in the full dataset regime and its y-axis position indicating its rank in the small dataset regime. The color of each point reflects the performance difference between the two dataset sizes.
Algorithms located along the diagonal line performed similarly in both the full and small dataset regimes. Examples of such algorithms based on F1 score in \cref{fig:rank-dff} include SSL \cite{159_kang2023cpath}, StainAug \cite{21_tellez2019cpath}, Macenko \cite{199_macenko2009cpath}, ANDMask \cite{parascandolo2020andmask}, EQRM \cite{eastwood2022eqrm}, and RSC \cite{huang2020rsc} among others.
Conversely, some algorithms showed improved performance rankings on smaller datasets, evident from their positions below the diagonal line. Notable examples include Transfer \cite{zhang2021transfer}, VREx \cite{krueger2021vrex}, and SagNet \cite{nam2021sagnet}. Ideally, algorithms that are versatile and perform consistently across different data conditions are desirable. Therefore, algorithms close to the diagonal line and lower left corner of plots in \cref{fig:rank-dff} (such as SSL and StainAug) are more suitable ones to investigate regardless of the size of the dataset at hand.

As mentioned before, in  CAMELYON17 and HISTOPANTUM, most algorithms perform well due to the large amount of data and simpler domain shifts. However, the level of performance for all algorithms drops on MIDOG22, which encompasses all kinds of DS. 
In the large-scale MIDOG22 dataset, SSL and StainAug showed the best performance in terms of F1 score, whereas in the small-scale sMIDOG22 dataset, SSL is not among the top-performing algorithms. Instead, Transfer \cite{zhang2021transfer} and SD \cite{pezeshki2021sd} algorithms achieve high F1 scores along with StainAug. Alternatively, focusing on the accuracy metrics obtained for the MIDOG22 dataset, we can see that the EQRM algorithm \cite{eastwood2022eqrm} is very promising. Although we have found that SSL and StainAug are generally good options to consider for DG applications, there is no one ``best'' algorithm that fits all the situations. Depending on the dataset size, the types of DS in the dataset, and the difficulty of the task different DG algorithms can perform differently. 

Notably, the baseline ERM algorithm \cite{vapnik1999erm} (which simply uses data from different domains in the mini-batches with standard data augmentation during training) performed consistently better than most DG algorithms, indicating that simple methods can be effective if implemented properly. This is in line with the findings of other works that investigated various DG algorithms \cite{gouk2022limitations,domainbed}. This underscores the importance of careful experiment design and incorporating well-performing baseline algorithms when looking for an optimal DG algorithm in any application.

\subsection{DG guidelines}
While we cannot pinpoint a single best algorithm for all scenarios, we can suggest general guidelines that help narrow down the number of DG algorithms to investigate for a specific application. First and foremost, ensure that the experiments are designed properly. For example, if cross-validation is implemented, make sure there is no data leakage and use domain-level stratification of cases between the train and test sets. Then, it is recommended to fine-tune a pretrained model (such as SSL \cite{159_kang2023cpath}), instead of learning from scratch or starting with ImageNet weights. Furthermore, implementing data augmentation as much as possible is recommended, especially modality-specific techniques such as stain augmentation (or at least HSV color augmentation) and generic data augmentation techniques (like image blurring, rotation, etc.). Incorporating augmentations has been shown to improve the generalizability of the model in many studies \cite{chlap2021review,faryna2024automatic,29_pohjonen2022cpath,94_kleppe2021cpath,59_alemi-koohbanani2020cpath,21_tellez2019cpath,196_jahanifar2022cpath,209_jahanifar2022cpath} by learning more robust feature representations that are invariant to confounding factors such as stain, sharpness, or rotations of the images.

After that, if desirable results are not achieved, a combination of the following algorithms can be investigated in conjunction with the suggested techniques above: ARM \cite{zhang2021arm}, CausIRL\_CORAL \cite{chevalley2022CausIRL}, Transfer \cite{zhang2021transfer}, and EQRM \cite{eastwood2022eqrm}. These algorithms have shown promise for DG and a high likelihood of convergence, and are less sensitive to parameter selection. On the other hand, based on our experience, investigation of IGA \cite{koyama2020iga} and SANDMask \cite{shahtalebi2021sand} can be ignored due to their sensitivity to parameter selection and low likelihood of convergence in CPath applications.
For more universal guidelines on how to achieve DG in CPath, please refer to \cite{jahanifar2023domain}.

\subsection{Limitations}
This study uses a ResNet50 model \cite{he2016deep} as the feature extractor with all DG algorithms. While ResNet50 is known for its robust performance in various tasks, the choice of model capacity and architecture can have a significant impact on the generalizability \cite{jiang2023domain,tan2019efficientnet}. A larger or alternative model architecture could potentially enhance performance. Our focus is specifically on the DG algorithm and the ResNet model due to its properties is considered to be well-suited for this study. Furthermore, one of the best-performing algorithms is SSL which is a ResNet50 model pretrained on TCGA image patches \cite{159_kang2023cpath}. However, there are many more recent foundation models released for CPath that have shown DG capability \cite{filiot2023scaling,vorontsov2024foundation,chen2024towards,campanella2024clinical}. We could not include any of these models as a DG algorithm in our comparison due to their considerable difference in model size and architecture from other algorithms, however, we expect that replacing the SSL \cite{159_kang2023cpath} with foundation models and also using them in conjunction with other recommended DG algorithms may further improve the results. This is an interesting question to investigate in future foundation model research areas. 

Furthermore, despite our efforts to determine the optimal working range for each algorithm, there remains the possibility that further tuning could yield improved performance. The results presented should be considered as initial findings from an out-of-the-box application of these methods. Future work could explore additional parameter adjustments to refine performance and potentially achieve better outcomes for some of the algorithms.

Admittedly, all of our benchmarking tasks are restricted to binary classification tasks. This choice is made to maintain a clear and focused evaluation framework, and also due to the limited availability of multi-labeled datasets. Moreover, this study does not explore other machine-learning tasks, such as semantic segmentation or bounding-box detection. This is partly because most DG algorithms have been tailored toward classification tasks. Although there are some DG algorithms suggested for other applications \cite{jahanifar2023domain}, it is hard to benchmark them mostly because there is no suitable platform or dataset in CPath to enable it. Nevertheless, investigating these additional tasks could reveal how the DG algorithms perform across a broader range of applications and potentially uncover different insights or challenges.

\section{Conclusions}
\label{sec:conclusion}
This study reports a comprehensive benchmarking of various domain generalization algorithms on three diverse computational pathology tasks. We evaluated 30 algorithms, including SOTA methods, self-supervised learning, and pathology-specific techniques, using a robust cross-validation framework. Our findings reveal that self-supervised learning and stain augmentation were consistently among the best-performing algorithms, emphasizing their effectiveness in addressing domain shifts. The baseline ERM algorithm also demonstrated competitive performance, highlighting the importance of careful experiment design and proper implementation of baseline algorithms.

The results underscore that no single ``best DG algorithm'' exists for all scenarios. The choice of the DG algorithm should be guided by factors like dataset size and diversity, domain shift types, and task complexity. Nevertheless, our study offers recommendations that may be useful for selecting effective DG strategies in CPath. We suggest fine-tuning a pretrained model, incorporating stain augmentation, and considering algorithms like ARM \cite{zhang2021arm}, CausIRL\_CORAL \cite{chevalley2022CausIRL}, Transfer \cite{zhang2021transfer}, and EQRM \cite{eastwood2022eqrm} for optimal performance.
We hope that this study will trigger further development of robust and generalizable deep learning models for CPath applications.

\label{sec:conc}




\section*{Acknowledgment}
During the writing of this manuscript, MJ collaborated with the BigPicture consortium (\url{www.bigpicture.eu}), and this project has received funding from the Innovative Medicines Initiative 2 Joint Undertaking under grant agreement No 945358. This Joint Undertaking receives support from the European Union’s Horizon 2020 research and innovation program and EFPIA (\url{www.imi.europe.eu}).

\appendix


 \bibliographystyle{elsarticle-num} 
 \bibliography{cas-refs}





\end{document}